\documentclass[runningheads]{llncs}
\usepackage[T1]{fontenc}
\usepackage{graphicx}
\usepackage{algorithm}
\usepackage{algorithmic}
\usepackage{amsmath}
\usepackage{subcaption}
\usepackage{float}

\begin{document}
\title{Memory-Augmented State Machine Prompting: A Novel LLM Agent Framework for Real-Time Strategy Games}
%
%
%
%
%
\author{Runnan Qi\inst{1}\orcidID{0009-0000-9572-4068} 
\and Yanan Ni\inst{1}\orcidID{0009-0008-9682-5504}
\and Lumin Jiang\inst{1}
\and Zongyuan Li\inst{2}
\and Kuihua Huang\inst{1}
\and Xian Guo\inst{2}
}

\authorrunning{Qi et al.}

\institute{
National University of Defense Technology, Changsha, China \\
\email{\{qirunnan13579, niyanan, khhuang\}@nudt.edu.cn}
\and
Nankai University, Tianjin, China \\
\email{\{2120230524, guoxian\}@mail.nankai.edu.cn}
}

\maketitle              
\begin{abstract}
This paper proposes Memory-Augmented State Machine \\Prompting (MASMP), a novel framework for LLM agents in real-time strategy games. Addressing key challenges like hallucinations and fragmented decision-making in existing approaches, MASMP integrates state machine prompting with memory mechanisms to unify structured actions with long-term tactical coherence. The framework features: (1) a natural language-driven state machine architecture that guides LLMs to emulate finite state machines and behavior trees through prompts, and (2) a lightweight memory module preserving strategic variables (e.g., tactics, priority units) across decision cycles. Experiments in StarCraft II demonstrate MASMP's 60\% win rate against the hardest built-in AI (Lv7), vastly outperforming baselines (0\%). Case studies reveal the method retains LLMs' semantic comprehension while resolving the "Knowing-Doing Gap" through strict state-action mapping, achieving both interpretability and FSM-like reliability. This work establishes a new paradigm for combining neural and symbolic AI in complex decision-making.

\end{abstract}
\section{Introduction}
Real-time strategy (RTS) games like \textit{StarCraft II} represent a grand challenge for AI, testing capabilities in real-time decision-making, long-term planning, and strategic adaptation. While reinforcement learning agents like \texttt{AlphaStar} have achieved superhuman performance\cite{ref_article1}, they require immense computational resources and lack interpretability. In contrast, Large Language Model (LLM)-based agents offer a promising alternative by simulating the human ``perception-reasoning-action'' cycle\cite{ref_url1}, demonstrating strong potential across domains from military planning (\texttt{COA-GPT}\cite{ref_proc1}) to complex game environments like \textit{Minecraft} (\texttt{GITM}\cite{ref_url2}).

However, in complex RTS environments, LLM agents face critical limitations that prevent them from competing effectively. They suffer from \textbf{hallucinations} (generating invalid actions), \textbf{greedy decision-making} (prioritizing short-term gains over long-term strategy), and \textbf{fragmented execution} (inconsistent actions across decision cycles due to a lack of memory). These issues result in poor performance; for instance, the \texttt{LLM-PySC2} agent achieves only an 8\% win rate against level-6 and 0\% against level-7 built-in AI.

To overcome these challenges, we propose the \textbf{Memory-Augmented State Machine Prompting (MASMP)} framework. Our work is built upon \texttt{LLM-PySC2}, a text-based API that provides a natural language interface for \textit{StarCraft II}, enabling LLMs to process game observations and output actions. MASMP integrates state-machine prompting to enforce structured, reliable decision-making and a strategic memory module to maintain long-term tactical coherence. Our agent achieves a 60\% win rate against the hardest built-in AI (Lv7), significantly outperforming all previous LLM-based baselines. This work demonstrates the potential of hybrid neuro-symbolic architectures for complex decision-making tasks.

\section{Related Works}

\subsection{Traditional RTS Game Agents}

Traditional RTS games have long relied on rule-based systems, where finite state machines (FSMs)\cite{ref_article2,ref_article3} and hierarchical FSMs\cite{ref_proc2} stand out for their simplicity and reliability. Behavior trees offer another effective approach, handling complex concurrent tasks through modular designs\cite{ref_article4}. These methods form the backbone of built-in AI systems in popular titles like \textit{StarCraft II} and \textit{Age of Empires}.

The field advanced significantly with reinforcement learning, particularly \texttt{AlphaStar}'s breakthrough in achieving superhuman performance in \textit{StarCraft II} through deep neural networks and imitation learning\cite{ref_article1}. Other approaches employing either RL or rule-based methods have also demonstrated strong performance against built-in AI\cite{ref_url3}.

However, both paradigms face inherent limitations: RL agents demand substantial computational resources and struggle to adapt to novel strategies, while rule-based systems require extensive manual engineering and lack true environmental comprehension.

\subsection{Large Language Models in RTS Games}

The emergence of LLM-based agents has introduced a new paradigm for RTS games. \texttt{TextStarCraftII}\cite{ref_article5} pioneered LLM integration with \textit{StarCraft II}, while \texttt{LLM-PySC2}\cite{ref_url4} enhanced this approach with multi-agent coordination and full action space support, establishing itself as a standard experimental platform.

This shift comes with significant challenges. LLM agents are plagued by hallucinations (generating impossible actions), local greediness (prioritizing short-term gains)\cite{ref_url5}, strategic inconsistency (incoherent planning across decision cycles), and a pronounced Knowing-Doing Gap (failing to execute well-reasoned plans)\cite{ref_url5}. Consequently, their performance remains limited, with reported win rates as low as 8\% against intermediate-level (Lv6) and 0\% against expert-level (Lv7) built-in AI\cite{ref_url4}.

Current improvement strategies include:
\begin{itemize}
\item \textbf{Prompt Engineering}: Employing few-shot learning and Chain-of-Thought in both \texttt{TextStarCraftII}\cite{ref_article4} and \texttt{LLM-PySC2}\cite{ref_url4}
\item \textbf{Hybrid Architectures}: Integrating rule-based automation (e.g., \texttt{Easy Build Mode}) in \texttt{LLM-PySC2}\cite{ref_url4} and combining LLM with FSM in \texttt{SwarmBrain}\cite{ref_url6}
\end{itemize}

These contributions highlight a crucial insight: \textbf{LLMs can benefit from traditional rule-based systems} as complementary modules for achieving stateful and reliable decision-making.

\section{Memory-Augmented State Machine Prompting}

\subsection{State Machine Prompting for LLM-Based Agents}

To enhance decision-making reliability in RTS games, we propose \textbf{State Machine Prompting}, a novel approach that guides LLMs to emulate structured decision patterns of finite state machines (FSMs) and behavior trees through natural language.

\begin{figure}[H]
\centering
\includegraphics[width=0.6\textwidth]{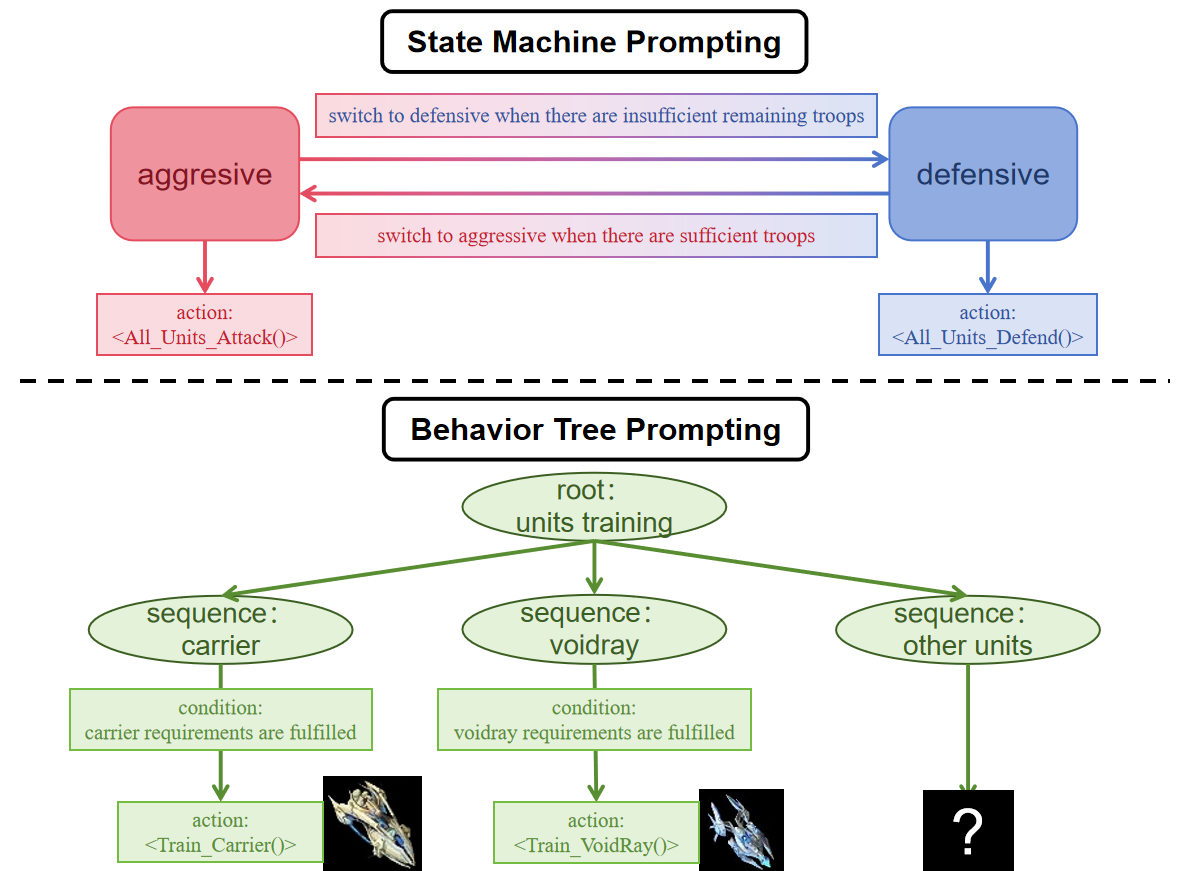}
\caption{Framework of State Machine Prompting for LLM-Based Agents.} \label{fig1}
\end{figure}

As shown in \textbf{Fig.1}, our framework comprises three key components:
\begin{itemize}
\item \textbf{Macro-Strategic State Machine}: Defines tactical states (e.g., \texttt{<aggressive>}), natural language transition conditions, and state-action mappings.
\item \textbf{Action Implementation Behavior Tree}: Implements hierarchical decision-making through selector, sequence, condition, and action nodes.
\item \textbf{Supplementary Atomic Rules}: Standalone natural language rules for specific scenarios.
\end{itemize}

Unlike traditional FSMs requiring exhaustive rule enumeration, our approach uses natural language conditions (e.g., "when resources exceed threshold"), leveraging LLMs' ability to generalize from partial specifications without manual edge-case handling.

\subsection{Strategic Memory for Non-Markovian Decision Making}

RTS games exhibit non-Markovian characteristics due to fog of war and strategic temporal dependencies. While prior works treated RTS as MDPs:
\begin{equation}
a_t \sim \text{LLM\_Generate}(o_t, \text{prompt})
\end{equation}

this assumption fails in practice. Our framework introduces strategic memory $M$ storing state variables (e.g., \verb|[Tactic]:<defensive>|), extending the formulation:
\begin{equation}
(s_t, a_t) \sim \text{LLM\_Generate}(o_t, M_{t-1}, \text{prompt}_{sm})
\end{equation}

\begin{equation}
M_t = \text{Update}(M_{t-1}, s_t)
\end{equation}

where $\text{prompt}_{sm}$ denotes our state machine prompt template, enabling persistent tactical coherence across decisions.

\subsection{Implementation in LLM-PySC2 Environment}

We implement our Memory-Augmented State Machine Prompting (MASMP) framework within \texttt{LLM-PySC2}, creating a closed-loop system for \textit{StarCraft II}.

\begin{figure}[H]
\centering
\includegraphics[width=0.8\textwidth]{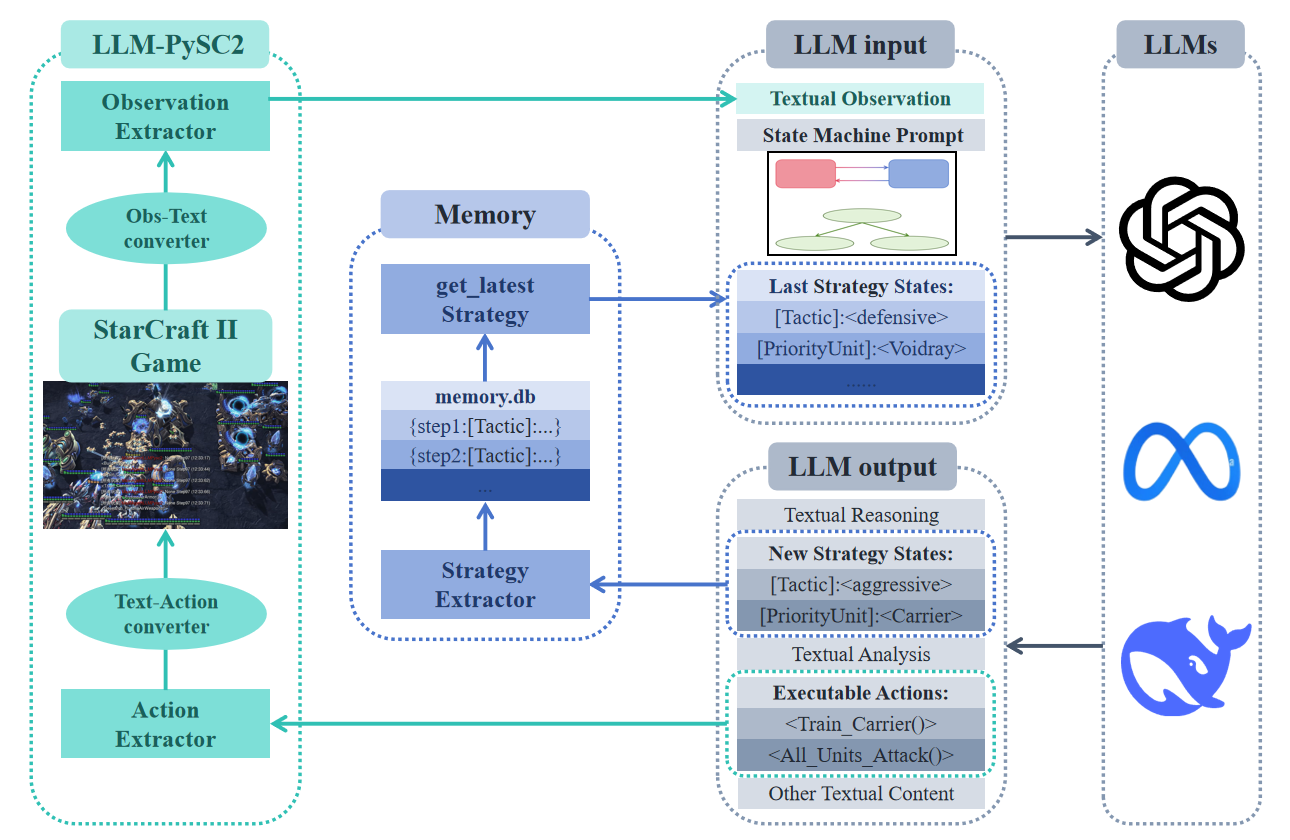}
\caption{MASMP framework within \texttt{LLM-PySC2}} \label{fig2}
\end{figure}

As shown in \textbf{Fig.2}, the system integrates textual observations with our prompt template and retrieved memory to form LLM input. The output is parsed for both action execution and strategy storage.
\begin{algorithm}[h]
\caption{Workflow of the MASMP Framework in \texttt{LLM-PySC2}}
\label{alg:masmp}
\begin{algorithmic}[1]
\REQUIRE Textual Game Observation $o_t$, MemoryDB $memory$, State Machine Prompt $prompt_{sm}$, Timestep $t$
\ENSURE Action execution, Memory update
\STATE $last\_strategy \gets memory.\text{get\_latest}()$ \COMMENT{Retrieve via MemoryDB method}
\STATE $input_t \gets \text{CONCAT}(o_t, prompt_{sm}, last\_strategy)$
\STATE $output \gets \text{LLM\_Generate}(input_t)$
\STATE $strategies \gets \text{StrategyExtractor.extract\_strategies}(output)$ \COMMENT{Regex pattern matching}
\IF{$strategies$ is not empty}
    \STATE $memory.\text{add\_memory}(strategies[0], t)$ \COMMENT{Store with timestep}
\ENDIF
\STATE $\text{ExecuteActions}(output)$
\end{algorithmic}
\end{algorithm}

\section{Experiments and Results}
\subsection{Experimental Setup}

We evaluated our method in the \texttt{LLM-PySC2} environment using \textit{StarCraft II}'s global gameplay scenario on map \texttt{Simple64}. Experiments used \texttt{DeepSeek-V3} with \texttt{Easy Build/Control Mode} enabled, testing against built-in AI (difficulty levels 1-7) under symmetric fair-play conditions. Win rate ($WR$) was used as the evaluation metric:

\begin{equation}
WR = \frac{N_{\text{win}}}{N_{\text{total}}} \times 100%
\end{equation}

\subsection{Experimental Results}

As shown in \textbf{Table.1}, our Memory-Augmented State Machine Prompting (MASMP) approach significantly outperforms the baseline across all difficulty levels. While the baseline fails completely at higher difficulties (0\% at Lv6/Lv7), MASMP maintains perfect win rates at Lv1-Lv5 and achieves remarkable 80\% and 60\% win rates at professional-grade Lv6 and Lv7 respectively. This demonstrates that LLM agents can now compete with professionally-engineered rule-based AI through our integrated approach.

\begin{table}[h]
\centering
\caption{Win Rate Comparison between Baseline and MASMP}
\label{tab:win_rates}
\begin{tabular}{lccccccc}
\hline
Method & Lv1 & Lv2 & Lv3 & Lv4 & Lv5 & Lv6 & Lv7 \\
\hline
Baseline & 100\% & 100\% & 100\% & 40\% & 40\% & 0\% & 0\% \\
MASMP & 100\% & 100\% & 100\% & 100\% & 100\% & 80\% & 60\% \\
\hline
\end{tabular}
\end{table}

\subsection{Comparative Analysis}

\subsubsection{Strategic Coherence \& Dynamic Comprehension}

\textbf{Fig.3} illustrates MASMP's dynamic strategy adaptation. The agent transitions from defensive to aggressive state upon achieving force advantage (Step75), maintains aggression while assessing battle progress (Step76), and strategically retreats when detecting reinforcements (Step77), successfully preserving forces (Step78). This demonstrates coherent tactical tempo control and causal reasoning capabilities absent in memoryless baselines.

\begin{figure}[htbp]
    \centering
    
    \begin{subfigure}[b]{0.48\textwidth}
        \centering
        \includegraphics[width=\linewidth]{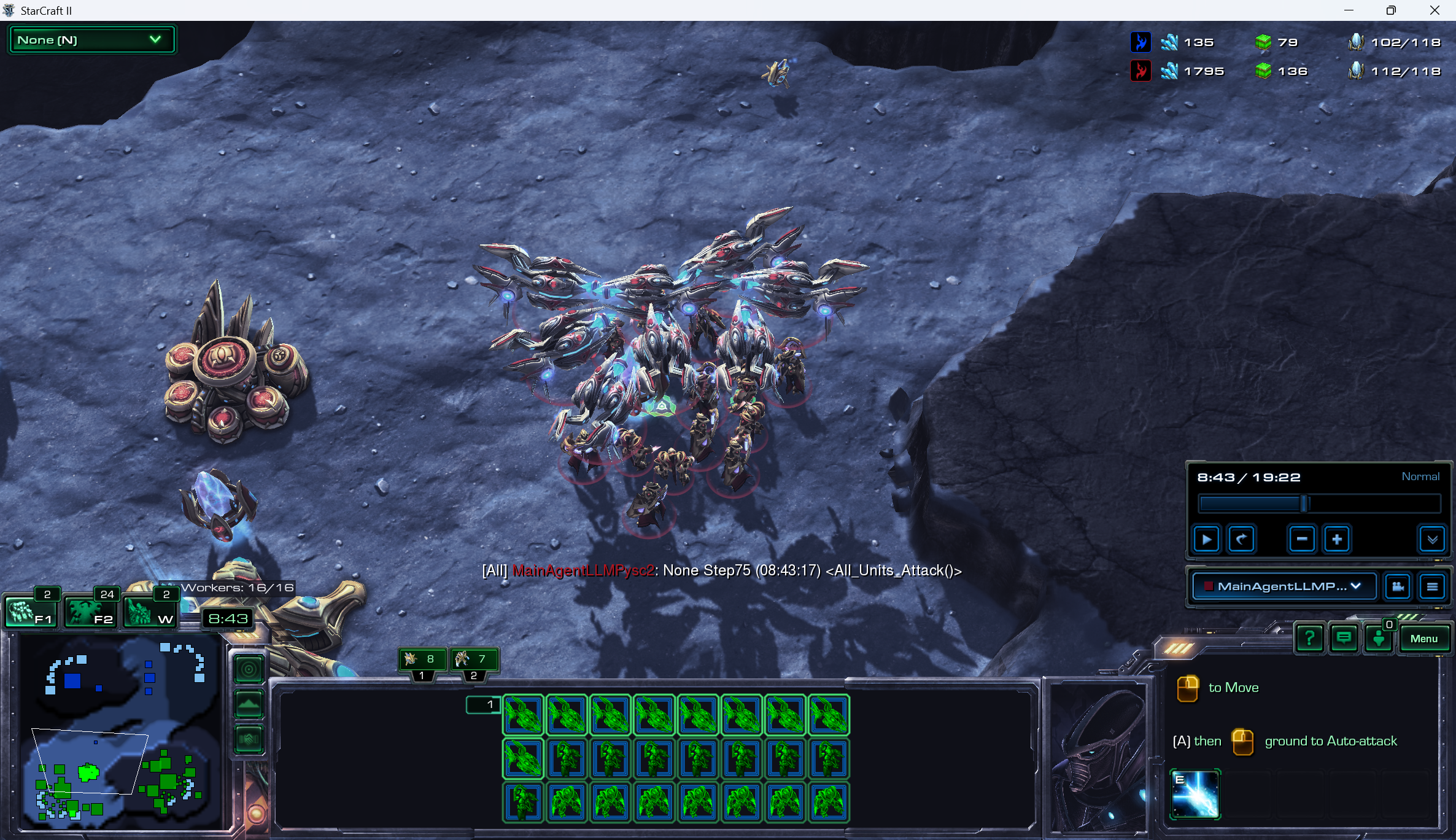}
        \caption{step75: \texttt{<defensive>} $\rightarrow$ \texttt{<aggressive>}}
        \label{fig:sub1}
    \end{subfigure}
    \hfill
    \begin{subfigure}[b]{0.48\textwidth}
        \centering
        \includegraphics[width=\linewidth]{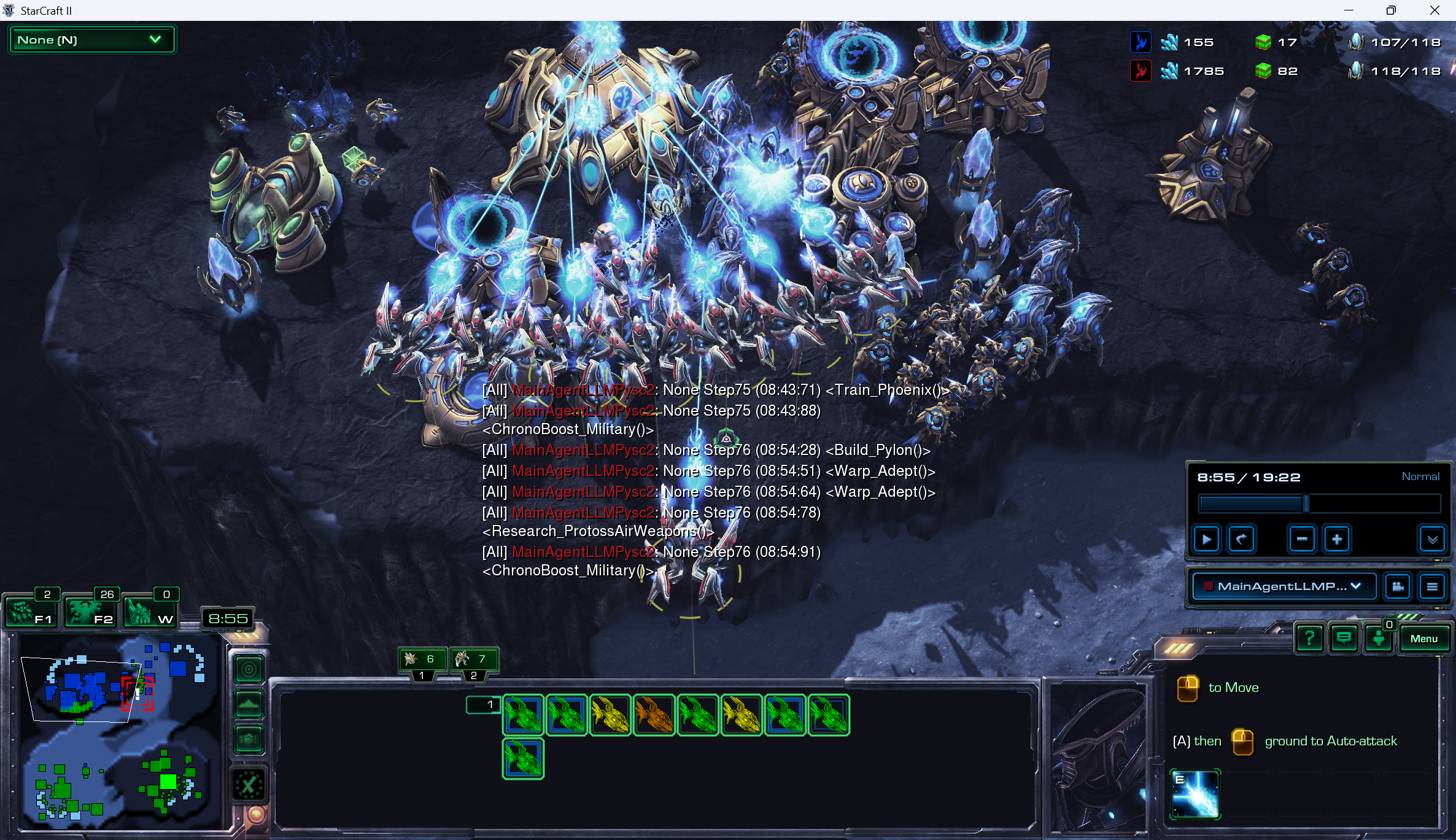}
        \caption{step76: \texttt{<aggressive>}}
        \label{fig:sub2}
    \end{subfigure}
    
    \vspace{0.2cm}
    
    \begin{subfigure}[b]{0.48\textwidth}
        \centering
        \includegraphics[width=\linewidth]{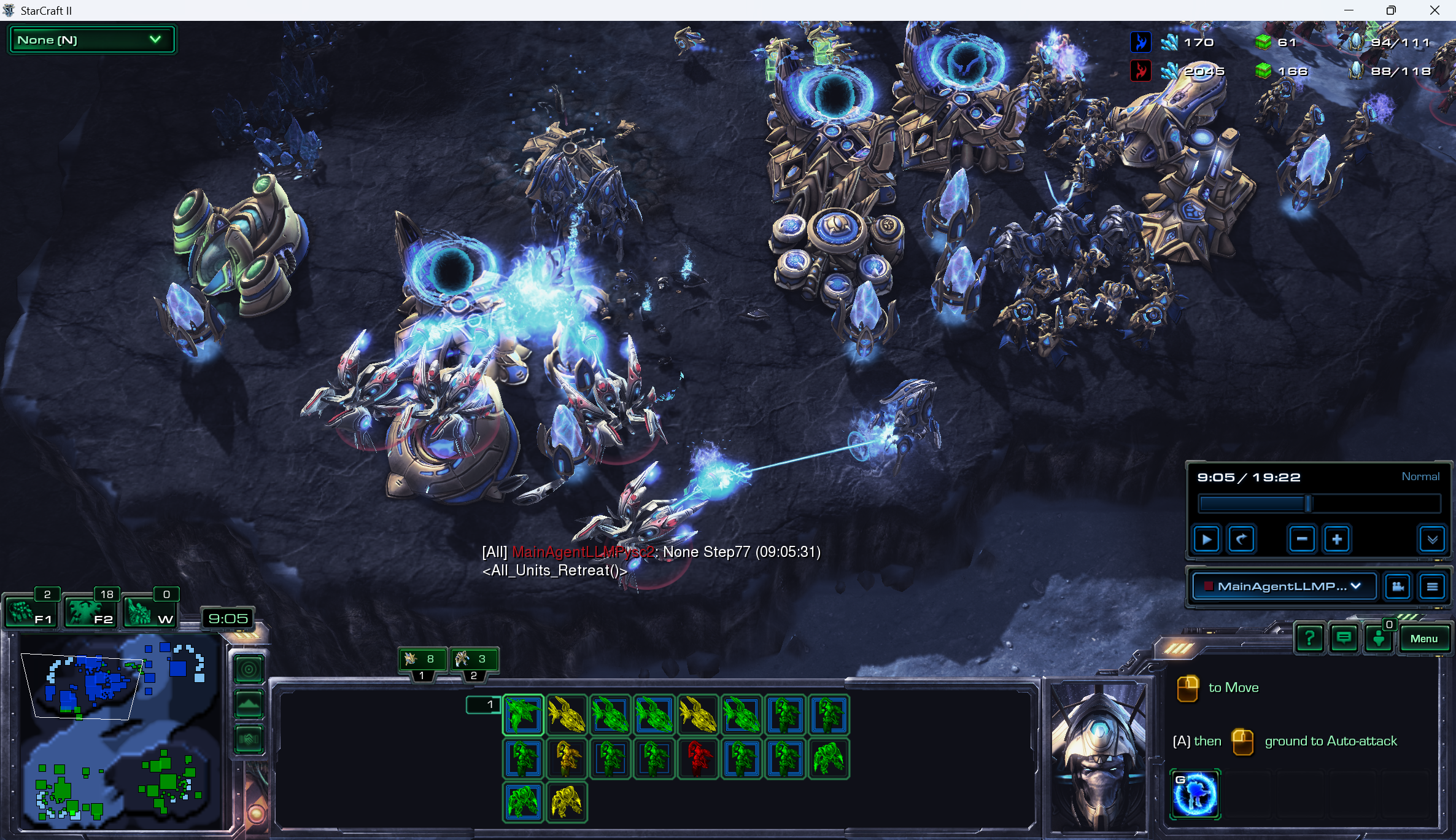}
        \caption{step77: \texttt{<aggressive>} $\rightarrow$ \texttt{<defensive>}}
        \label{fig:sub3}
    \end{subfigure}
    \hfill
    \begin{subfigure}[b]{0.48\textwidth}
        \centering
        \includegraphics[width=\linewidth]{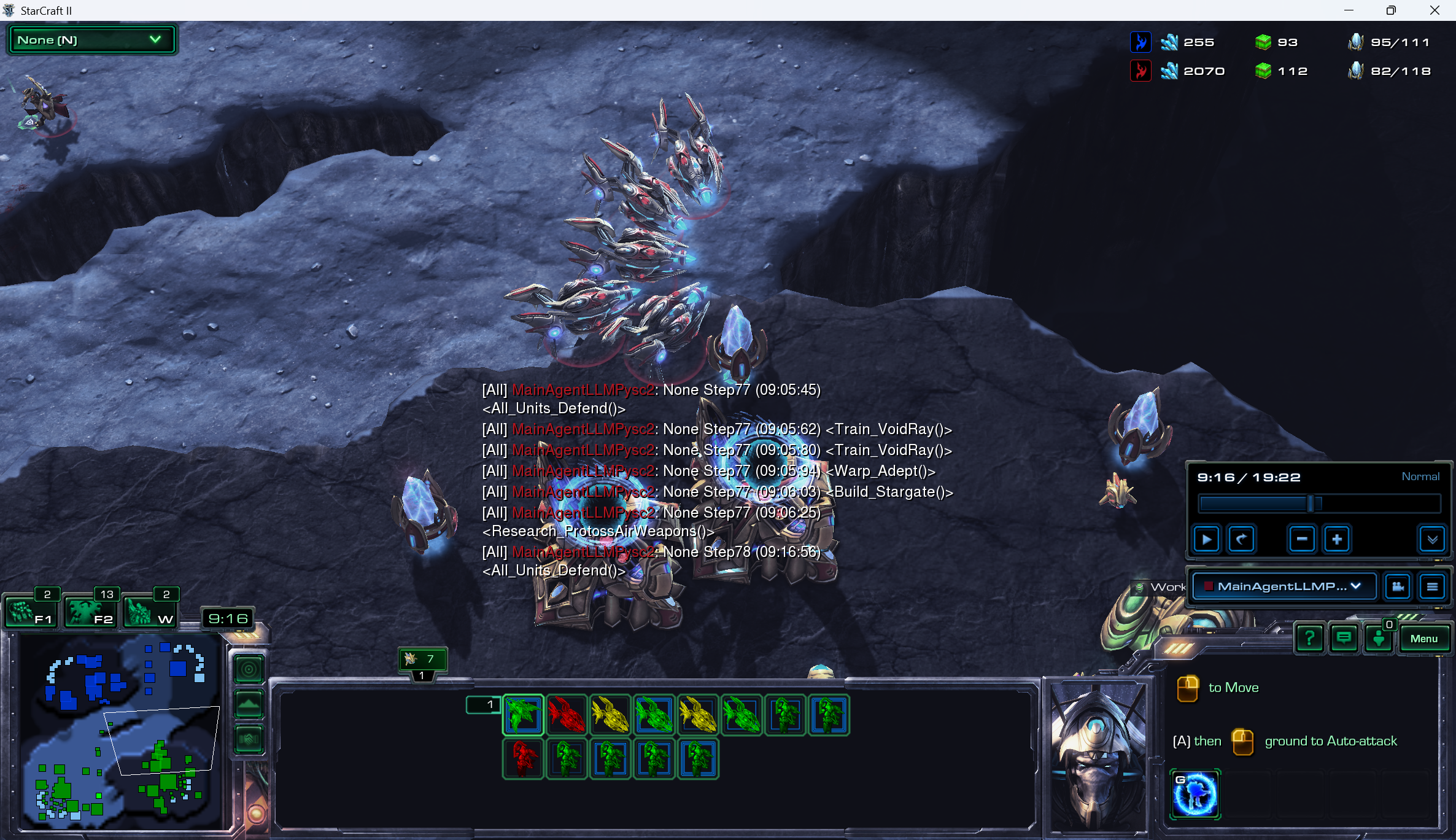}
        \caption{step78: \texttt{<defensive>}}
        \label{fig:sub4}
    \end{subfigure}
    
    \caption{Dynamic Strategy Adaptation Case}
    \label{fig:total}
\end{figure}

\subsubsection{Solving the Greedy Trap: Long-Term Planning}

\begin{figure}[h]
\centering
\includegraphics[width=0.8\linewidth]{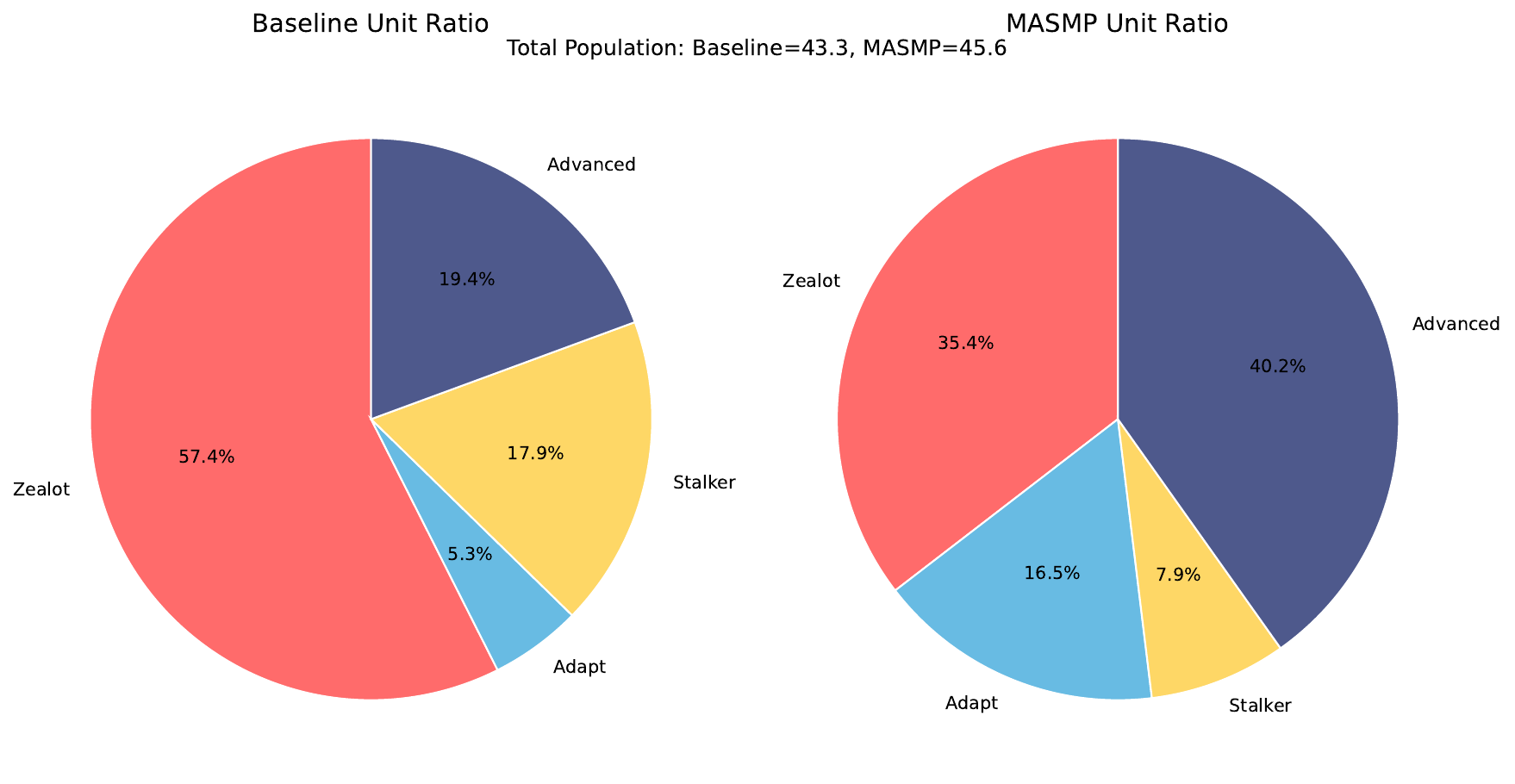}
\caption{Early-game Unit Production Ratio}
\label{fig:unit_production}
\end{figure}

\textbf{Fig.4} quantitatively shows MASMP's superior long-term planning. At 7-minute mark, MASMP produces 18.32 advanced units (40.2\% of total) versus baseline's 8.40 (19.6\%), with better diversification (Zealot ratio: 35.5\% vs. 57.8\%). Through state variables like \texttt{[PriorityUnit]}, our method guides resource allocation toward technological advancement, avoiding the baseline's greedy trap of spamming low-tier units.

\subsubsection{Advantages Over Traditional State Machines}

MASMP maintains structural constraints while preserving LLMs' advantages:
\begin{itemize}
\item \textbf{Interpretability}: Natural language justifications for state transitions
\item \textbf{Generalization}: Semantic adaptation to unseen scenarios
\item \textbf{Creativity}: Autonomous employment of unspecified counters
\end{itemize}

The probabilistic formulation enables fuzzy reasoning that eliminates traditional FSMs' need for precise thresholds and manual rule programming.

\section{Conclusion and Future Work}

We proposed \textbf{Memory-Augmented State Machine Prompting (MASMP)}, a novel framework that bridges LLM flexibility with rule-based reliability for RTS games. MASMP integrates state machine prompting with strategic memory, achieving a 60\% win rate against \textit{StarCraft II}'s highest-difficulty AI (Lv7), significantly outperforming baselines (0\%). This demonstrates the potential of hybrid neuro-symbolic architectures for complex decision-making. Future work includes exploring multi-agent coordination, dynamic prompt optimization, and cross-domain applications.

%
%
%
%

\end{document}